\documentclass[utf8]{FrontiersinHarvard} 

\usepackage{url,hyperref,lineno,microtype,subcaption}
\usepackage[onehalfspacing]{setspace}
\usepackage{todonotes}
\usepackage{booktabs}

\newcommand{\rev}[1]{\textcolor{black}{#1}}

\def\keyFont{\fontsize{8}{11}\helveticabold }
\def\firstAuthorLast{Kowald {et~al.}} 
\def\Authors{Dominik Kowald$^{1,2,*}$, Sebastian Scher$^{1,4}$, Viktoria Pammer-Schindler$^{1,2}$, Peter Müllner$^{1}$, Kerstin Waxnegger$^{1}$, Lea Demelius$^{1,2}$, Angela Fessl$^{1,2}$, Maximilian Toller$^{1}$, Inti Gabriel Mendoza Estrada$^{1}$, Ilija \v{S}imi\'{c}$^{1}$, Vedran Sabol$^{1}$, Andreas Trügler$^{1,2,4}$, Eduardo Veas$^{1,2}$, Roman Kern$^{1,2}$, Tomislav Nad$^{3}$, Simone Kopeinik$^{1,*}$}

\def\Address{$^{1}$Know-Center GmbH, Graz, Austria \\
$^{2}$Graz University of Technology, Graz, Austria \\
$^{3}$SGS Digital Trusts Services GmbH, Graz, Austria\\
$^{4}$University of Graz, Graz, Austria}

\def\corrAuthor{Dominik Kowald, Simone Kopeinik}

\def\corrEmail{dominik.kowald@tugraz.at, skopeinik@know-center.at}

\begin{document}
\onecolumn
\firstpage{1}

\title[Establishing and Evaluating Trustworthy AI: Overview and Research Challenges]{Establishing and Evaluating Trustworthy AI: Overview and Research Challenges} 

\author[\firstAuthorLast ]{\Authors} 
\address{} 
\correspondance{} 

\extraAuth{}

\maketitle

\begin{abstract}
Artificial intelligence (AI) technologies (re-)shape modern life, driving innovation in a wide range of sectors. However, some AI systems have yielded unexpected or undesirable outcomes or have been used in questionable manners. As a result, there has been a surge in public and academic discussions about aspects that AI systems must fulfill to be considered trustworthy. In this paper, we synthesize existing conceptualizations of trustworthy AI along six requirements: 1) human agency and oversight, 2) fairness and non-discrimination, 3) transparency and explainability, 4) robustness and accuracy, 5) privacy and security, and 6) accountability. For each one, we provide a definition, describe how it can be established and evaluated, and discuss requirement-specific research challenges. 
Finally, we conclude this analysis by identifying overarching research challenges across the requirements with respect to 1) interdisciplinary research, 2) conceptual clarity, 3) context-dependency, 4) dynamics in evolving systems, and 5) investigations in real-world contexts. 
Thus, this paper synthesizes and consolidates a wide-ranging and active discussion currently taking place in various academic sub-communities and public forums. It aims to serve as a reference for a broad audience and as a basis for future research directions. 

\tiny
 \keyFont{\section{Keywords:} trustworthy ai, artificial intelligence, fairness, human agency, robustness, privacy, accountability, transparency}
\end{abstract}

\section{Introduction}
\label{s:intro}

From sophisticated chatbots like Chat-GPT to AI-driven recommender systems enhancing our entertainment experiences on platforms like Netflix and Spotify~\citep{anderson2020algorithmic}, the impact of AI on our lives is significant. 
AI-based decision support systems are proving invaluable in critical fields such as life science and healthcare~\citep{rajpurkar2022ai}. Similarly, AI is reshaping hiring and human resources practice ~\citep{van2021machine} and transforming the banking and finance landscape with innovative solutions~\citep{cao2022ai}. 
However, in the past, some AI systems have been used in questionable manners, which has led to unexpected or undesirable results. Examples include biased algorithms perpetuating discrimination in recruitment processes~\citep{chen2023ethics} or AI-driven recommender systems favoring popular content and, with this, users interested in popular content~\citep{kowald2020unfairness,kowald2022popularity}. 
Alongside biases in algorithms, AI systems rely on training data, including personal and private user information, which raises concerns for potential privacy and security breaches. One example is the Equifax data breach, in which private data records of millions of users were compromised~\citep{zou2018concern}. Additionally, when thinking of self-driving cars, unreliable AI-based systems could even cause physical harm, as demonstrated by the unfortunate Uber car crash in 2018, in which a malfunctioning algorithm did not detect and, as a consequence, killed a pedestrian on the road~\citep{kohli2020enabling}. 

As a consequence, there has been an increase in public and academic discussions about the essential requirements AI systems must fulfill to be considered trustworthy. There is also a growing consensus on the necessity of setting up standards and regulations to ensure and validate the trustworthiness of AI. In this respect, the European Commission (EC) has proposed the AI Act~\citep{madiega2021artificial}, a comprehensive regulatory framework for supporting the responsible development and deployment of AI technologies within the European Union. The AI Act seeks to establish clear rules governing the development and deployment of AI systems while imposing strict requirements for high-risk AI applications. The various interpretations of trustworthy AI add further complexity to this discourse by encompassing not just technical requirements but also human-centered and legal considerations. 
\rev{Another important framework proposed by the European Commission has been the ``Assessment List for Trustworthy AI (ALTAI)''~\citep{ALTAI,radclyffe2023assessment}, which enables organizations to self-assess the trustworthiness of AI solutions based on a checklist.} 

\begin{figure}[t!]
\begin{center}
\includegraphics[width=.80\textwidth]{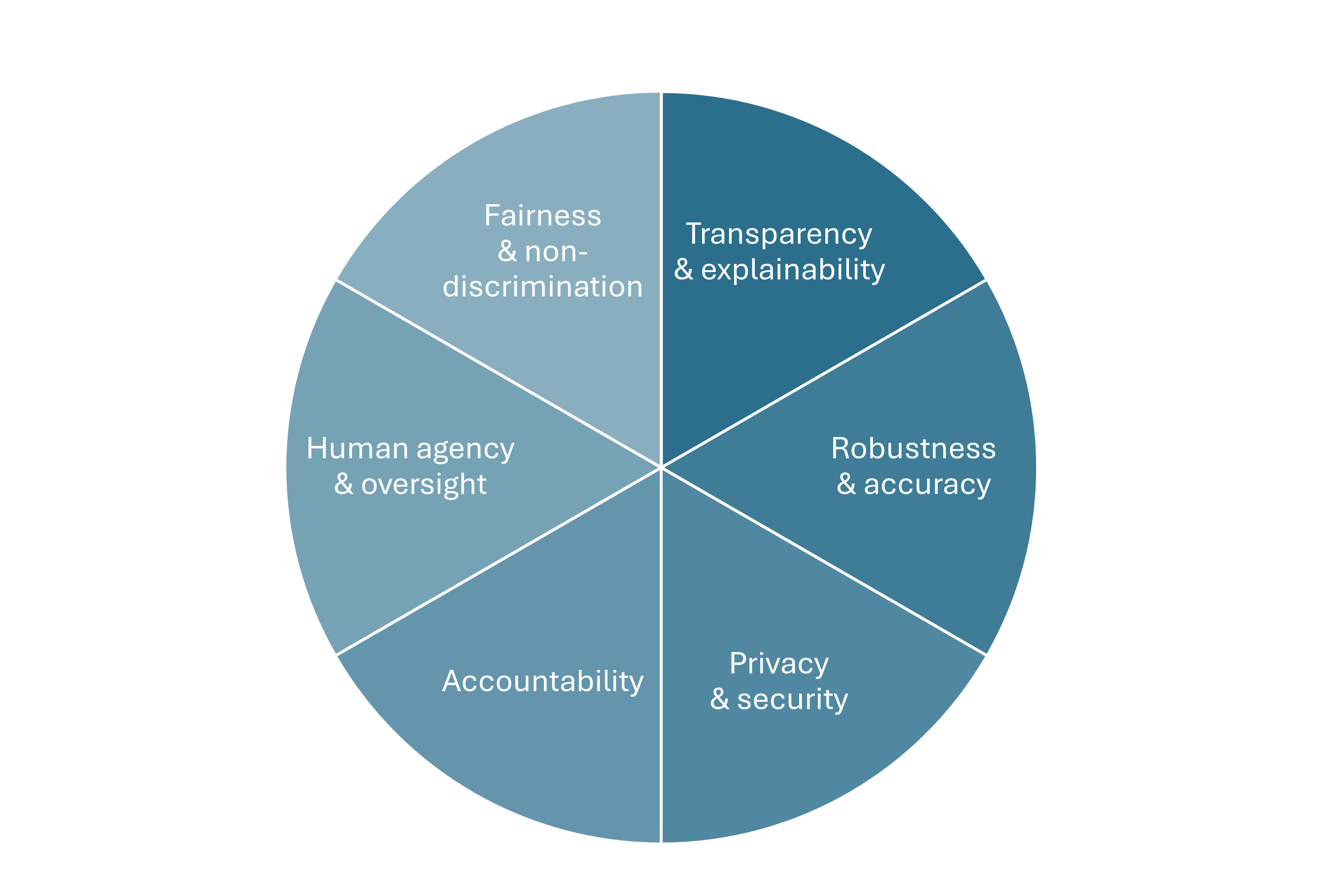}
\caption{An illustration of the six requirements of trustworthy AI investigated in this paper.}
\label{f:intro_requirements}
\end{center}
\end{figure}

This paper contributes insights into this discourse by analyzing the state-of-the-art regarding six aspects of AI systems that are typically understood as requirements for systems to be viewed as trustworthy. These requirements are: 1) human agency and oversight, 2) fairness and non-discrimination, 3) transparency and explainability, 4) robustness and accuracy, 5) privacy and security, and 6) accountability (see Figure~\ref{f:intro_requirements}). We define each of these six requirements, introduce methods to establish and implement these requirements in AI systems, and discuss corresponding validation methods and evaluation metrics. Such validation efforts are crucial from scientific and practical perspectives and might serve as a prerequisite for certifying AI systems and models~\cite {winter2021trusted}. Finally, for each of these requirements, we outline ongoing research challenges and future research perspectives. 

The contributions of our work are two-fold: firstly, we give a comprehensive overview of the requirements of trustworthy AI, in which we cover different viewpoints on trustworthy AI, including technical and also human-centered and legal considerations. Secondly, we discuss open issues and challenges in defining, establishing, and evaluating these requirements of trustworthy AI. 
\rev{Therefore, the guiding research question of this work is defined as follows: \textit{What is the current state of research regarding the establishment and evaluation of comprehensive - technical, human-centered, and legal - requirements of trustworthy AI?} To address this research question, we follow the methodology described in Section~\ref{sec:methodology}}. 

\rev{Our work complements existing surveys and articles on trustworthy AI in two main ways. Firstly, existing overview articles such as~\citep{chatila2021trustworthy,thiebes2021trustworthy,akbar2024trustworthy,diaz2023connecting} tend to focus on definitions of trustworthy AI and neglect evaluation aspects, which is one key aspect of our article. 
Specifically, related surveys such as~\citep{liang2022advances,wing2021trustworthy,emaminejad2022trustworthy} focus on specific aspects of trustworthy AI implementation and evaluation, namely data, formal methods, and robotics, respectively. In contrast, our article aims to provide a domain- and method-independent overview of trustworthy AI, which reflects the whole AI-lifecycle, including the evaluation phase.}  
\rev{Secondly, concerning validation and evaluation schemes for trustworthy AI, existing technical conceptualizations of trustworthy AI such as~\citep{floridi2021establishing,kaur2022trustworthy,li2023trustworthy} have focused on technical and reliability-oriented requirements such as transparency, privacy, and robustness. In contrast, in our paper, we discuss methods and open challenges towards establishing and evaluating trustworthy AI also through the lens of human-centric and legal requirements such as fairness, accountability, and human agency. Therefore, to the best of our knowledge, our paper is the first to investigate all six requirements of trustworthy AI in a unified way by discussing implementation and evaluation aspects across the whole lifecycle of trustworthy AI and outlining open research challenges and issues for all six requirements.}  

Our article shows that while evaluation and validation methodologies for technical requirements like robustness may rely on established metrics and testing procedures (e.g., for model accuracy), the assessment of human-centric considerations often requires more nuanced approaches that consider ethical, legal, and cultural factors. As such, our article emphasizes the need for further research to develop robust evaluation schemes that can be applied in research and practice across a variety of AI systems, particularly in high-risk domains where human values and rights are at stake (e.g., healthcare).

Next, in Section~\ref{sec:background}, we describe the relevant background for this article, including general definitions of AI and its lifecycle, and introduce the six requirements of trustworthy AI covered herein. After discussing each requirement separately in Section~\ref{s:requirements}, the paper closes with a conclusion and an outlook into future research directions in Section~\ref{sec:conclusion}.  

\section{Background}
\label{sec:background}
In this section, we give a short overview of definitions and preliminaries relevant to our article, introduce the six requirements of trustworthy AI discussed, \rev{and describe the methodology of our investigation}. 

\subsection{Definitions and Preliminaries of Trustworthy AI}
For our understanding of AI in the context of this work, we adhere to the definition outlined in the EU AI Act (adopted text, Art 3(1)\footnote{\url{https://www.europarl.europa.eu/doceo/document/TA-9-2024-0138_EN.pdf}}), in which AI is defined as ``\textit{a machine-based system designed to operate with varying levels of autonomy and that may exhibit adaptiveness after deployment and that, for explicit or implicit objectives, infers, from the input it receives, how to generate outputs such as predictions, content, recommendations, or decisions that can influence physical or virtual environments.}'' 
This definition encompasses a broad spectrum of algorithmic implementations, from simple logistic regression models to complex machine-learning approaches. In the article at hand, we consider this spectrum of AI systems, recognizing the diverse challenges and requirements that are associated with ensuring its trustworthiness. 

\begin{figure}[t!]
\begin{center}
\includegraphics[width=.60\textwidth]{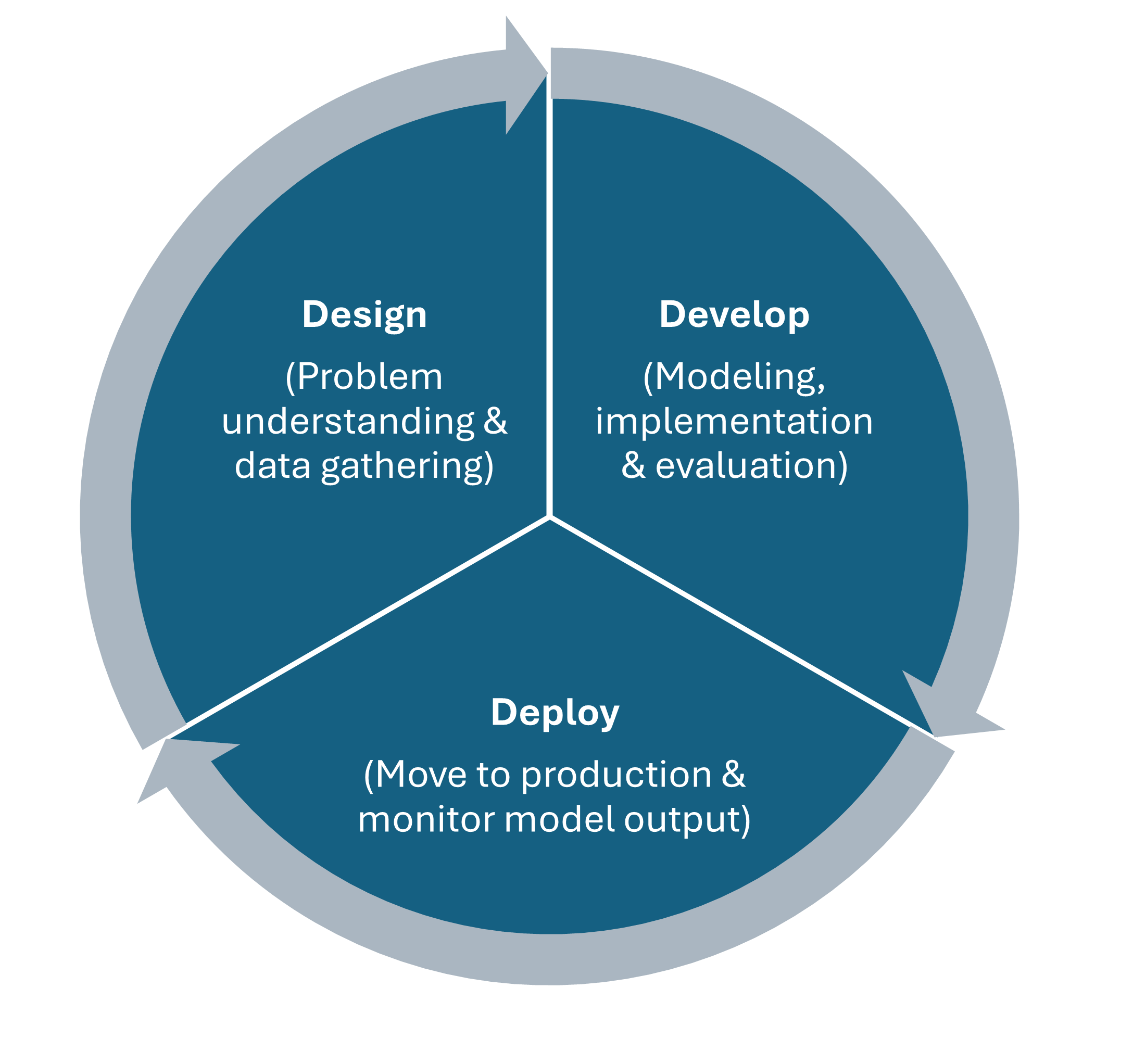}
    \caption{The AI-lifecycle. 
    The trustworthiness of AI can be conflicted in all phases - the design phase, the development phase, and the deployment phase.}
    \label{fig:AILifecycle}
\end{center}
\end{figure}

Additionally, we aim to consider the trustworthiness of AI from a holistic perspective that can be influenced in all phases of the AI-lifecycle, as thoroughly described in ~\citet{haakman2021ai} (see Figure~\ref{fig:AILifecycle}). These phases encompass the design, development, and deployment of AI-based systems and their designated tasks. While discussing the requirements of trustworthy AI, we refer to the phases where needed in this article.  
To comply with trustworthy AI, special attention should be paid to considering AI requirements in the \textit{design phase} throughout requirements engineering, problem understanding, and data collection strategies, the \textit{development phase}, comprising model implementation (e.g., optimizing feature weights), documentation, and evaluation, and finally, the \textit{deployment phase} including the integration of the AI model into a production environment, and the continuous monitoring and updating of the model. 

\subsection{Requirements of Trustworthy AI}
\label{sec:trustai_requirements}
Over the last years, various frameworks, guidelines, survey articles, and collections of requirements of trustworthy AI have been developed and published by researchers, governments, and private organizations~\citep{smuha2019eu,kaur2021requirements,floridi2021establishing,kaur2022trustworthy,li2023trustworthy,yeung2020recommendation}. Although these investigations differ with respect to the exact wordings, they agree on four fundamental principles that need to be considered when developing and validating trustworthy AI: 1) respect for human autonomy, 2) fairness, 3) explicability, and 4) prevention of harm~\citep{smuha2019eu,kaur2021requirements,kaur2022trustworthy}. In the following, we describe six requirements of trustworthy AI that are manifested within these principles. 

Hereinafter, principle 1 (respect for human control) is mainly associated with \textbf{human agency and oversight} (requirement 1, see Section~\ref{s:human_agency}), which refers to sustaining the autonomy of humans affected by AI systems, given different levels of human-AI interaction. The second principle (fairness) aims for the equal treatment of all affected individuals and subpopulations (i.e., defined by age, gender, education, ...). \textbf{Fairness and non-discrimination} (requirement 2, see Section~\ref{s:fairness}) describes the absence of bias in AI decisions that could result in unfair, unequal treatment that negatively affects certain people. Next, principle 3 (explicability) ensures the AI system is transparent and explainable. In particular, \textbf{transparency and explainability} (requirement 3, see Section~\ref{s:transparency}) is defined as the understandability of an AI system and the provision of information to explain the AI model's decisions.

\begin{figure}[t!]
    \centering
    \includegraphics[width=0.9\linewidth]{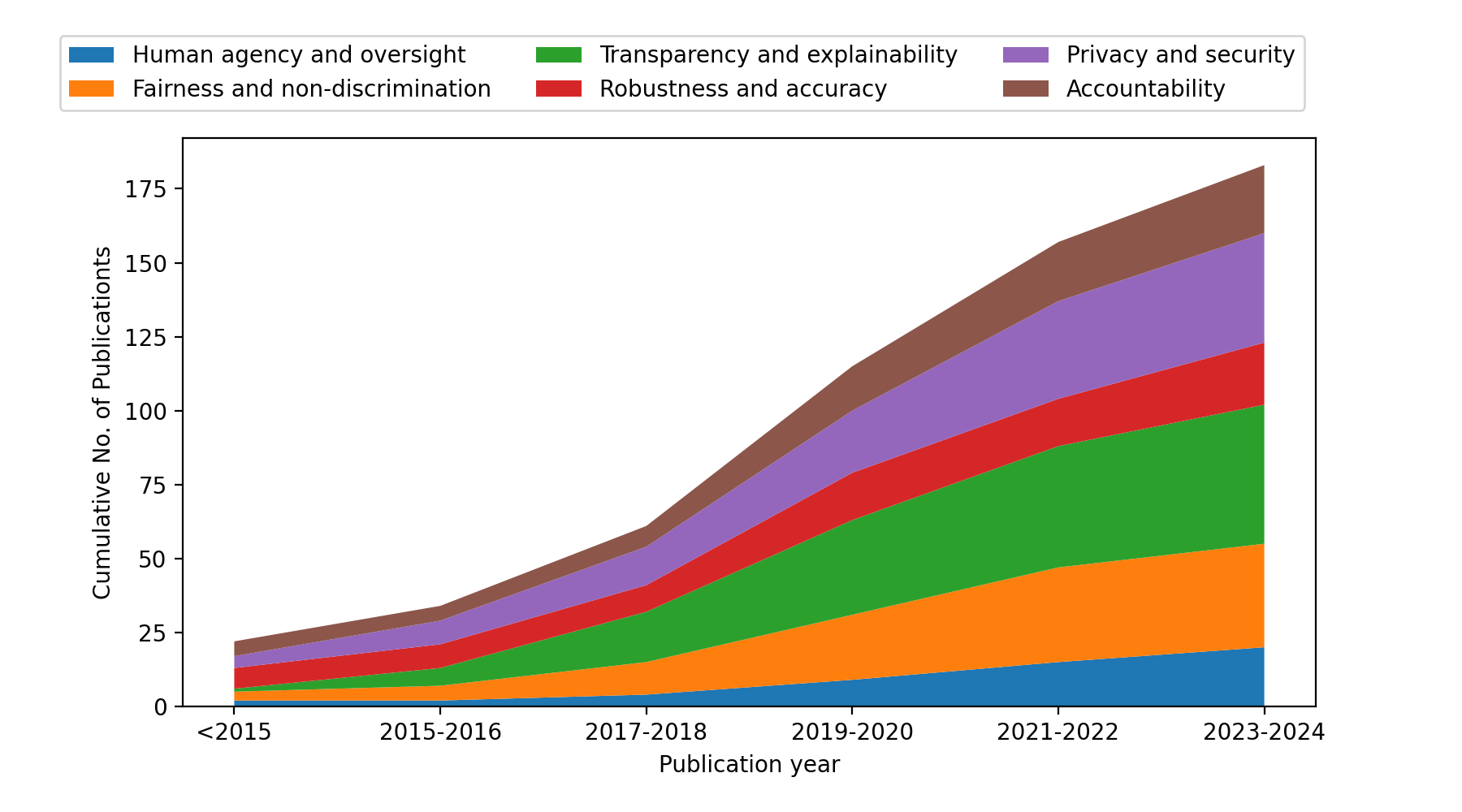}
    \caption{\rev{The number of publications per requirement included in this paper across publication years. We investigate 183 publications: 20 publications for \emph{human agency and oversight}, 35 publications for \emph{fairness and non-discrimination}, 47 publications for \emph{Transparency and explainability}, 21 publications for \emph{robustness and accuracy}, 37 publications for \emph{privacy and security}, and 23 publications for \emph{accountability}.}}
    \label{fig:enter-label}
\end{figure}

Finally, principle 4 (prevention of harm) should ensure that the AI system does not cause any harm to humans, society, or the environment. \rev{For example, it should be prevented that AI-based systems harm or, even worse, kill humans, which unfortunately has happened, e.g., in the aforementioned Uber car crash in 2018, in which a pedestrian was killed by a malfunctioned self-driving car~\citep{kohli2020enabling}.} This principle includes a range of requirements, including technical and legal concerns. An essential technical requirement within this principle is \textbf{robustness and accuracy} (requirement 4, see Section~\ref{s:robustness}), which is related to the performance of AI models and their ability to function in unexpected circumstances. 
Additionally, the principle ``prevention of harm'' is linked to \textbf{privacy and security} (requirement 5, see Section~\ref{s:privacy}) that focuses on protecting the personal and sensitive information of users in AI systems and on preventing malicious attacks on AI models. Concerning legal aspects, \textbf{accountability} (requirement 6, see Section~\ref{s:accountability}) entails the understanding of who is responsible for the decisions of AI systems and to ensure that mechanisms are in place to interfere with negative consequences. 
The literature (e.g.,~\citep{kaur2021requirements,smuha2019eu}) also discusses other important requirements of trustworthy AI within this principle, such as safety, reproducibility, sustainability, societal and environmental wellbeing, and data governance. 
While we highlight these requirements' significance, we believe they serve as overarching aspects that underpin the six other requirements of trustworthy AI discussed in this article. However, we strongly suggest a foundational commitment to safety, reproducibility, sustainability, societal and environmental well-being, and data governance when developing and validating trustworthy AI. In the next section, we discuss in detail these six requirements outlined above. 

\subsection{\rev{Methodology}}
\label{sec:methodology}
\rev{To collect relevant resources, we conduct an exploratory approach to define the research field and, for the main part, follow a semi-structured literature review \citep{snyder2019literature}. This allows the consideration of interdisciplinary literature to 1) specify a comprehensive set of aspects of trustworthiness in AI, 2) synthesize available knowledge regarding these aspects that is relevant when aiming to design, implement, and evaluate trustworthy AI, and 3) identify open challenges and knowledge gaps in these regards. Thus, we conducted exploratory literature research on trustworthy AI in general, which resulted in the six herein-discussed requirements. Then, the following procedure was completed for each requirement: 1) a Scopus search for conference papers, articles, and reviews, 2) the screening of the 100 most relevant abstracts as ranked by Scopus, 3) the screening of the remaining papers, and the extraction of relevant content, 4) snowballing and additional search in Google Scholar to close information gaps.} 

\rev{The goal of this paper was not to cover all existing publications of the field but rather to generate a comprehensive understanding of relevant research directions and their existing challenges. Therefore, we excluded articles with over-specialization, such as solutions only applicable to specific use cases or domains, and articles with limited contributions. This resulted in a collection of 183 papers as illustrated in Figure~\ref{fig:enter-label}.}

\section{Overview and Discussion of Trustworthy AI Requirements} \label{s:requirements}
In the following, we discuss six requirements an AI-based system should meet to be considered trustworthy. Each requirement is first defined, then we describe methods to establish and evaluate it, and finally, we debate open issues and research challenges. 
\rev{Table~\ref{tab:def} provides a complementary illustration of the problem definitions of these six requirements.}

\begin{table}[t!]
\centering
\caption{\rev{Problem definitions of the six trustworthy AI requirements investigated in this paper.}}
 \label{tab:def}
\begin{tabular}{l l}
\toprule
\rev{\textbf{Trustworthy AI requirement}} & \rev{\textbf{Problem definition}}\\ \midrule
\rev{Human agency and oversight} & \rev{Sustaining the autonomy of humans affected by AI systems}\\
\rev{Fairness and non-discrimination} & \rev{Mitigating bias in AI decisions to prevent unfair treatment}\\
\rev{Transparency and explainability} & \rev{Improving the understandability of an AI system and its decisions}\\
\rev{Robustness and accuracy} & \rev{Sustaining the model's performance (in unexpected circumstances)}\\
\rev{Privacy and security} & \rev{Protecting personal information of users in AI systems}\\
\rev{Accountaibility} & \rev{Understanding who is responsible for the decisions of AI systems} \\ \bottomrule
\end{tabular}
\end{table} 

\subsection{Requirement 1: Human Agency and Oversight} \label{s:human_agency}

\subsubsection{Definition of Human Agency and Oversight}
The principle of human agency and oversight refers to the idea that AI systems should uphold individual autonomy and dignity, and need to operate in a way that allows for substantial human control over the AI system's impact on people and society. This principle further postulates that AI systems should contribute to a democratic, flourishing, and equitable society and allow for human supervision to foster fundamental rights and ethical norms~\citep{ec2019ethics}. 

Although the terms \textit{human agency} and \textit{human oversight} are very alike and sometimes used as synonymous, they are not interchangeable~\citep{bennett2023does}. In this paper, we understand the term and concept \emph{human agency} as referring to the very broad idea that humans as intentional actors should be in control, particularly with respect to substantial and important parts of their lives~\citep{ec2019ethics}. AI systems shall not restrict this agency; rather, it would be desirable that through AI systems, human agency is increased. AI systems could, for instance, limit human agency by deceiving or manipulating users. However, users should be able to influence automated decisions and to fairly evaluate or question the AI system. Consequently, users who are impacted by AI systems or who oversee AI systems need to be able to acquire or to be equipped with related competencies and skills (AI Literacy) to understand and engage with AI systems to a satisfying degree~\citep{long2020ai,pammer-schindler-2022-AI-literacy}.

The term and concept of \emph{human oversight} is more specifically related to how AI systems are used and suggests that AI systems do not operate entirely autonomously, but instead, humans should oversee the way AI systems ``work'' within a more extensive process. This concept, therefore, is concerned with forms of human-AI interaction or collaboration, postulating that humans should be in a supervisory and decision-making role. Human oversight activities include observing, interpreting, and interfering in AI operations. Human oversight can be understood as a specific approach to facilitating human agency.

\citet{long2020ai} define AI literacy as \textit{"a set of competencies that enables individuals to evaluate AI technologies critically; communicate and collaborate effectively with AI; and use AI as a tool online, at home, and in the workplace."} Being AI literate means having a basic understanding of AI that empowers users to better interact with AI systems as they are able to judge the outcomes provided and, at the same time, to retain autonomy and agency~\citep{hermann2022artificial}. Facilitating AI literacy on a large scale is currently a subject both of research~\citep{ng2021conceptualizing} and of public endeavor. 
 
\subsubsection{Methods to Establish Human Agency and Oversight} 
Very broadly speaking, socio-technical and human-centric design methods~\citep{baxter-2010-ST-systems-design-engineering} are approaches for designing (AI-based) systems that can systematically consider human users and people impacted, from the very early stage of designing the systems~\citep{dennerlein-2020-ST-reflection-ethical-design-in-TEL}. 
The consideration of human factors in the design phase of the AI-lifecycle (see Figure~\ref{fig:AILifecycle}) responds well to the complex dynamics of the issues. Humans can interact with AI in various ways, which may require different levels of human agency and oversight. \citet{anders2022finding} propose to think of different patterns of human engagement with AI-based operations and decisions as follows\rev{, sorted along the decreasing involvement of humans}: 
\rev{
\begin{itemize}
\item \emph{Human-in-command}: Humans manage and oversee an AI system's overall operation, including its wider impact on society, economy, law, and ethics. Decisions at a high level are made on when and how to use an AI system~\citep{anderson2022human}. 
For example, regulatory bodies set ethical guidelines for the deployment of AI systems in healthcare to ensure patient safety and data privacy.
\item \emph{Human-in-the-loop}: Humans can intervene in AI-based decisions as well as in different steps of the underlying (typically machine learning based) algorithms~\citep{mosqueira2023human,munro2021human}. In medical diagnostics, for example, an AI system can suggest potential diagnoses, but a physician reviews and decides on the diagnosis and treatment plan. Note that often, this kind of human-computer collaboration is not possible or even desired~\citep{ec2019ethics}.
\item \emph{Human-on-the-loop}: Humans can intervene through (re-)designing the AI system or through involvement in system operations, where their role is primarily on monitoring~\citep{anderson2022human}. For example, a human operator may remotely monitor the performance of an autonomous vehicle and intervene only in emergency situations.
\item \emph{Human-out-of-the-loop}: Humans do not intervene. This could mean allowing an AI-based system to work without human involvement for specific tasks or in a completely automated manner. For example, in fully automated sections of vehicle production assembly lines, car parts can be assembled without human interaction. 
\end{itemize}
}
Other authors categorize the spectrum of involvement in the design and operations of AI-based systems differently. For instance,~\citet{fanni2020active} suggests a distinction between active and passive agency. Passive agency occurs when there are limited or no communication features that provide explanations for the decisions made by the AI system, or when users are uninformed about the potential consequences of AI interventions. This passive agency relates to the concepts of human-out-of-the-loop and human-on-the-loop. In contrast, active agency refers to situations where humans play a critical role in the design and operations of the AI-based system. This relates to human-in-the-loop and human-in-command approaches. 
\citet{WangHumanAgency2023} proposes the level of involvement to be defined by decisions and actions undertaken by humans and by AI-based systems. The author provides an example pattern of human-AI interaction as ``AI Suggests, Human Intelligence (HI) Decides'', which can be interpreted such that the AI is providing recommendations, but humans are taking the role of final decision makers. Overall, thinking about such patterns of human-AI interaction allows deciding - at the time of designing and using an AI-based system - what kind of interaction is desirable or possible. 
Finally, we introduce the notion of AI literacy as positively contributing to human agency and oversight. This applies to both the users of AI-based systems and the decision-makers responsible for regulating and deciding which systems are used and how or what AI-related competencies users need to deal with AI systems. 

AI literacy can be obtained in two dominant ways. First, through education in AI, particularly about everyday activities and technology~\citep{zimmerman2018teaching}. Such education can also be mediated by technology.
For instance, researchers are engaging young learners in creative programming activities, including AI~\citep{kahn2017child,zimmermann2019youth}.
Secondly, everyday AI systems could be designed to support users in being or becoming AI literate. \citet{long2020ai} outlined 15 concrete design considerations to promote users' understanding and learning when interacting with AI systems. For example, AI systems could provide visualizations and explanations of decision-making processes to enhance users' comprehension. They could also offer users the opportunity to learn about the system's reasoning processes by putting themselves “in the agent’s shoes”; to encourage users to investigate the used data in terms of source, data collection processes, and known limitations or encourage users \textit{"to be critical consumers of AI technologies by questioning their intelligence and trustworthiness"}.
Such support of AI literacy by AI systems is precious, as complex knowledge is highly context- and activity-dependent, and transferring knowledge from one context or activity to another can be quite challenging~\citep{eraut2004informal}.

Overall, systems need to be designed to be understandable for humans~\citep{long2020ai,ng2021conceptualizing}. This relates to the long-standing concepts of usability and learnability of systems. Finally, some of the above-described concepts, such as explaining decision-making, are essential for supporting transparency and explainability (see Section~\ref{s:transparency}).

\subsubsection{Methods to Evaluate Human Agency and Oversight}
Building upon the previous discussion of human agency, human oversight, and AI literacy and their interrelations, we propose considering evaluation as moving upward the hierarchy of dependencies: 

\begin{itemize}
\item \emph{AI Literacy:} AI literacy of relevant stakeholders is considered a prerequisite for human agency and oversight. This level can be assessed, for instance, through knowledge or competency tests, by checking certifications and formal degrees, or by investigating educational opportunities that relevant stakeholders have accessed or utilized.
\item \emph{System Understandability:} It is critical to evaluate whether the AI-based system or functionality communicates understandably how it operates and what effects user actions might have. This relates to established concepts of usability and learnability and newer, AI-specific concepts like transparency and explainability (see Section~\ref{s:transparency}). 
\item \emph{Human oversight:} It needs to be established whether and how the intended interaction pattern of human oversight is present in the evaluation setting. This evaluation level concerns observing the designed activity, with a focus on establishing whether humans are, reasonably and in an engaged manner, involved in the process, either in-the-loop, on-the-loop, or in-command. 
\item \emph{Human agency:} This is probably the most challenging concept to verify. One could argue that the existence and evaluation of a human-centric and socio-technical (AI) design process implies a certain level of human agency. The discussions led during this process could provide insights into how human agency is conceptualized and implemented in the AI-based system.  In the inverse, it may be possible to establish its absence, i.e., when it becomes noticeable that human agency is limited through an AI-based system or functionality.
\end{itemize}

\subsubsection{Open Issues and Research Challenges}

Challenges concern the \emph{conceptualization of human agency, oversight, and AI literacy} as interwoven concepts, \rev{and the operationalization in design patterns of (interfaces for) AI.} \rev{Such developments will need to be made} in relation to maturing technology \rev{such as increasing shop-floor automation at the conjunction of Internet of Things and AI-enabled data analytics, or the usage of generative AI across many sectors of knowledge-based and creative work}. Additionally, more and better synthesized \emph{design-oriented knowledge that captures how to design for human agency, oversight, and AI literacy} is pending. To date, research \rev{is revisiting the value of these design principles, e.g., in the context of AI~\citep{long2020ai} or generative AI~\citep{simkute-2024-ironies-generative-AI,weisz-2024-design-principles-genAI}}. 
However, \rev{these design principles need clearer examples of how to operationalize them concretely within applications.} 
Finally, due to the unclear conceptualization and the plethora of different designs with little grouping into overarching design patterns, \emph{the evaluation of human agency, oversight, and AI literacy} will remain challenging. Evaluations will also need to uncover how these concepts interact with design patterns, ethics, and trust in AI systems~\citep{ec2019ethics}, actual decisions made in the domain of interest, and the overall socio-technical system performance (i.e., how good are decisions in the broader context and for whom). 

\subsection{Requirement 2: Fairness and Non-Discrimination} \label{s:fairness}

\subsubsection{Definition of Fairness and Non-Discrimination}
As AI products are being increasingly used in various fields and domains, their influence and impact on society are discussed not only in the machine learning community (e.g.,~\citet{8825881}) but also among the general public. AI may negatively impact individuals and society by reproducing existing societal stereotypes that can adversely affect vulnerable groups~\citep{dubal2023algorithmic}. The unjust treatment of specific populations or individuals is particularly concerning in sensitive fields such as criminal justice, employment, education, or health, as it can result in significant consequences such as being refused medical care~\citep{seyyed2021underdiagnosis} or educational opportunity~\citep{chang2021targeted}. Previous instances of such misconduct have been documented, including Google Ads showing lower-paid jobs to women and minority groups\footnote{\url{https://incidentdatabase.ai/cite/19/\#r184}}, Apple Card granting lower credit limits to women than equally qualified men\footnote{\url{https://incidentdatabase.ai/cite/92/\#r2037}}, and commercial facial recognition systems performing poorly for women with black skin \citep{pmlr-v81-buolamwini18a}. 

People's perception of fairness strongly depends on the context, which can include various factors, such as socio-political views, personal preferences, or the particular context and use case \citep{saxena2019perceptions}. 
Especially in AI systems, achieving fairness is a multifaceted problem.
Algorithmic fairness describes the absence of bias in AI decisions that would favor or disadvantage a person or group in a way that is considered unfair in the context of the application \citep{ntoutsi2020bias,srivastava2019mathematical}.
Bias, often also called ``discriminatory'' or ``unfair'' bias, refers to outcomes of disproportionate advantage or disadvantage for a specific group of individuals, i.e., ``\textit{systematic discrimination combined with an unfair outcome is considered bias}'' \citep{bird2019fairness}. 
Consequently, we refer to fairness as the absence of discriminatory or ``unfair'' bias towards individuals, items, or groups. 

Although ethical concerns are often at the forefront of public discourse, ``unfair'' bias can significantly impact society and businesses, even in seemingly non-critical domains. Therefore, it is crucial to consider the various risks from a business perspective. According to a report by \citet{Deloitte2021}, biased AI bears risk for several negative consequences. These include missing out on potential business opportunities, damaging reputation, and facing regulatory and compliance issues.
One example of missing out on opportunities is when a recommender system only benefits a particular user group~\citep{kowald2022popularity}. While the members of the advantaged group may find the system useful, other groups don't experience the same level of system performance and stop using the product. This results in a loss of potential customers for the platform provider.
Another consequence of biased AI is reputational damage, especially when the technology fails to address sensitive societal issues. For instance, using face recognition software that only works well for parts of the ethnicities in the user population will likely lead to negative public perception and backlash against the company.
Finally, in cases where anti-discrimination laws govern the use of AI, such as in the job market, unfair algorithms can lead to legal problems. For example, an HR system that discriminates based on gender, age, or race can result in fines and penalties for the company. More information on accountability can be found in Section~\ref{s:accountability}.

Furthermore, the issue of bias and fairness is complex because bias is naturally inherent in human behavior~\citep{doi:10.1177/1745691619855638} and thus, ``unfair'' bias can be introduced in every stage of the AI-lifecycle, as illustrated in Section~\ref{sec:background}. This problem becomes even more challenging in evolving AI systems because they can potentially reinforce bias between the user population, data, and algorithm \citep{baeza2018bias}. Thus, monitoring and addressing ``unfair'' bias throughout the entire AI-lifecycle is essential. 

\subsubsection{Methods to Establish Fairness and Non-Discrimination}
A wide range of methods has been proposed to increase fairness in AI models~\citep{bellamy2019ai,barocas2021fairness}. Because fairness strongly depends on the context, making AI models fair means making them fair in a particular context, i.e., according to an appropriate definition of fairness~\citep{srivastava2019mathematical}. 
Depending on their application level in the AI-lifecycle, ``bias mitigation approaches'' or ``fairness enhancing methods'', are commonly grouped into three categories~\citep{bellamy2019ai,pessach2023algorithmic,barocas-hardt-narayanan}:
\begin{enumerate}
    \item \emph{Pre-processing} concerns improving the training data's quality and balancing its composition in regard to protected groups. This can be applied independent of the AI algorithm. Pre-processing can regulate fairness in acceptance rates but does not cater to other fairness constraints. Examples of pre-processing algorithms include reweighting \citep{calders2009building}, optimized preprocessing \citep{calmon2017optimized}, learning fair representations \citep{zemel2013learning}, and disparate impact removal \citep{feldman2015certifying}. 
    \item \emph{In-processing} describes the design and optimization of an AI algorithm towards an explicitly defined, fair solution. Thus, it incorporated fairness in the training algorithms themselves and can only be applied to specific algorithms/models that are well understood~\citep{srivastava2019mathematical}. 
    \item \emph{Post-processing} aims to adapt the AI model's results towards a balanced distribution for protected groups. Examples thereof include methods for calibration, constraint optimization, or setting thresholds for the maximum accuracy differences between groups~\citep{pleiss2017fairness}. 
\end{enumerate}

While each approach has its particular pros and cons, all of them potentially negatively affect the models' accuracy (see Section~\ref{s:robustness}).

\subsubsection{Methods to Evaluate Fairness and Non-Discrimination}
The auditing or evaluation of algorithmic fairness can, similar to the mitigation strategies, be approached according to the three main phases of the AI-lifecycle, i.e., design, development, and deployment \citep{koshiyama2021towards}. Examples of what can be assessed are 1) population balance and fair representation in data (design), 2) the implementation of fairness constraints in modeling or the adherence to fairness metrics in evaluation (development), and 3) the adherence to fairness metrics in real-time monitoring (deployment) \citep{akula2021audit}.

Measures of algorithmic bias are a quantitative evaluation of the result set of the system at hand \citep{pessach2023algorithmic}. 
The highest level of separation between different definitions of fairness is between individual and group fairness, which are related to the legal concepts of disparate treatment and disparate impact, respectively \citep{barocas2016big}.  
\begin{itemize}
    \item \emph{Individual fairness} considers fairness on an individual level and requires treating similar individuals similarly.
    \item \emph{Group fairness} calculates fairness on a group level, requiring different groups to be treated equally. 
 \end{itemize}   
Furthermore, one can differentiate between three principal approaches: fairness in acceptance rates, fairness in error rates, and fairness in outcome frequency \citep{barocas2021fairness}. 
\citet{verma2018fairness} provides an overview of the 20 most prominent definitions. One challenge, however, is selecting the ``right'' definition and metrics, as many different definitions of algorithmic fairness and related metrics exist. In many settings, these definitions contradict each other – thus, it is usually not possible for an AI model to be fair in all three aspects. The appropriate metrics must be selected for a given application and its particularities. 
Several software packages are available that implement important metrics. Popular open-source frameworks include AIF 360~\citep{bellamy2019ai}, Fairlearn~\citep{bird2020fairlearn}, and Aequitas~\citep{saleiro2018aequitas}.

\subsubsection{Open Issues and Research Challenges}
Fairness is a concept highly context-dependent that, in practice, may require ethical consultation \citep{john2022some}. No fairness definition applies to all contexts, and it seems an intrinsic—and unsolvable—challenge of the field to formulate legally compliant measurements mathematically \citep{wachter2021fairness}. 
What is perceived as fair or unfair also varies between different cultural and legal settings. It remains unclear how to tune the fairness of an AI application intended to be used in multiple cultural or legal contexts~\citep{srivastava2019mathematical} and, more generally, how to apply and assess existing regulations, standards, and ethical constraints in practice~\citep{costanza2022audits}.

From a more technical perspective, ensuring fairness when combining different AI components poses a significant challenge. This can be particularly difficult when reusing AI tools or algorithms with limited access to code, or when exchanging data audited only for a specific use case or application context. In fact, it has been shown that measures of algorithmic fairness are sensitive to any alterations in the input data and to even simple changes in train-test splits \citep{friedler2019comparative}. In principle, monitoring fairness in AI systems that are in production is possible (e.g., \citet{vasudevan2020lift}). However, it is still much more demanding to define when fairness criteria are met and when not because the algorithm's performance may change over time (e.g., \citet{lazer2014parable}). 
\rev{The application of generative AI models presents additional challenges, especially in the context of language. Despite a significant body of research, it is still unclear how to effectively measure and evaluate their bias and how to transform these measurements to be suitable for application in various contexts or to consider different minority characteristics \citep{nemani2023gender}.}

Finally, there is no standard to determine the adequate trade-off between different fairness metrics nor between fairness and accuracy.
While there have been attempts to show that the fairness-accuracy trade-off is rather an issue of historical bias in data \citep{dutta2020there}, it remains unclear how to generate an ideal, unbiased dataset as a standard in practice. 

\subsection{Requirement 3: Transparency and Explainability} \label{s:transparency}

\subsubsection{Definition of Transparency and Explainability}
Transparency and explainability are two related but distinct concepts. Explainability aims to enhance comprehension, build trust, and facilitate decision-making \citep{adadi2018peeking}. 
In contrast, the goal of transparency is to ensure understandability and accountability~\citep{lepri2018fair,arias2022focus,mcdermid2021artificial}. 
With transparent and explainable AI, users can better estimate the trustworthiness of AI systems since they can understand their inner workings and, consequently, their opportunities and limitations \citep{naiseh2023different}. 
However, there is no final consensus on the definition and scope of transparency to date. Therefore, in this paper, we will focus on explainability as a means to achieve transparency in AI models. 

Furthermore, including explainability methods in an AI system offers additional benefits, such as allowing for debugging and expert verification of the AI system, which can foster task accuracy and efficiency \citep{weber2023beyond} \rev{(e.g. in healthcare~\citep{albahri2023systematic,hulsen2023explainable})}. 
For example, \citet{anders2022finding} improved a model's prediction accuracy by utilizing explainability methods to identify dataset samples that led the model to learn spurious correlations. 
\citet{young2019deep} used explainable AI to help experts verify that models for melanoma detection relied on the correct data aspects for their predictions. Other researchers have developed explainability methods to facilitate iterative learning and improvement of AI models by identifying patterns, biases, or errors in the model's decision-making processes~\citep{ribeiro2016should,mehdiyev2020explainable}.  

\subsubsection{Methods to Establish Transparency and Explainability}
Methods to make AI systems explainable are often summarized under the umbrella term ``Explainable AI (XAI)''~\citep{holzinger2022explainable}, and sometimes also interpretable AI \citep{molnar2020interpretable}. XAI aims to increase transparency and explainability in AI systems to ensure trust, understanding, and accountability. 
With respect to the AI-lifecycle (see Section~\ref{s:requirements}), XAI is relevant in all three phases: in the design phase, for understanding and incorporating stakeholders' explainability requirements; in the development phase, for understanding important data aspects, performing error analysis and model refinement; and finally, in the deployment phase for continuous model verification and enhancing user trust. 
XAI methods vary in their approach towards achieving these goals and can be categorized based on their \textit{scope} and \textit{model dependence}. 
The term \emph{scope} refers to whether XAI methods can produce global or local explanations \citep{samek2021explaining}. 
Global explanations aim to explain the AI model as a whole and, thus, provide insight into the model's general decision-making process, e.g., SHAP \citep{lundberg2017unified}. In contrast, local explanations focus on the model's decision-making process in regard to a single sample, making the explanation more specific (e.g., LIME \citep{ribeiro2016should}). 
Global explanations tend to be computationally more expensive, as they need to consider the entire training input space, whereas local explanations might work on one input sample. 

Furthermore, XAI methods can be classified as ``model-agnostic" or ``model-specific" methods \citep{arrieta2020explainable}, according to their \emph{model dependence}. 
Model-agnostic XAI methods can provide explainability and transparency for any AI system. 
In contrast, model-specific XAI methods are tailored to a single AI architecture, which may limit their compatibility with other AI systems. However, they tend to create more accurate and translucent (i.e., the extent to which the explanations rely on particularities of the inner workings of the AI system) explanations than model-agnostic methods~\citep{carvalho2019machine}. Additionally, model-specific XAI methods can not be implemented for every AI system. Meanwhile, model-agnostic XAI methods can be implemented on every AI system, as they often use the AI system as an oracle. They do so by probing the model many times to estimate the effects of the input on the model prediction, which can lead to expensive computations. 
Examples of model-agnostic XAI methods are SHAP \citep{lundberg2017unified}, LIME \citep{ribeiro2016should} and the broader category of counterfactual explanations \citep{guidotti2022counterfactual}; model-specific XAI methods are DeepLIFT \citep{shrikumar2017learning} and Integrated Gradients \citep{sundararajan2017axiomatic}. 

It is a common belief that ``white-box'' models - models whose ``inner workings'' can be inspected - are immediately interpretable and transparent. 
However, they often have lower prediction accuracy than more complex models~\citep{moreira2022benchmarking}. 
In addition, white-box models, e.g., Linear Regression and Decision Trees, often need extra steps to be used or treated as ``full-fledged'' XAI methods because in order to be most effective, XAI methods are required to be understandable and not overwhelming to their target users, meaning that they are specifically tailored to meet their requirements~\citep{miller2017explainable}. 
Social science research regards high-quality explanations as a form of conversation and proposes explanation theories like temporal causality, social constructivism, and attribution theory~\citep{mendoza2023importance}. 

Deep learning models have succeeded across various domains by utilizing computational units called neurons, ordered in sequential layers forming neural network (NN) models. 
NNs can autonomously learn meaningful internal features without manual feature engineering~\citep{DeepLearning2015}. Consequently, to train the models, we often use raw data directly or include all features, regardless of complexity~\citep{Roy2015}. However, while NNs map inputs directly to outcomes, they do not disclose how features are weighted in relation to the model's output~\citep{Zhao2015HFS}.

XAI methods can provide explanations in many different ways. 
\emph{Feature attribution methods} generate values for each input feature, highlighting its importance to the AI model's predictions.  
However, these methods can be sensitive to input noise and correlated features, resulting in misleading conclusions \citep{NEURIPS2018_294a8ed2}. 
They are commonly presented textually (numerically), via bar charts \citep{ribeiro2016should}, or via heatmaps \citep{sundararajan2017axiomatic}. 
XAI methods can also provide explanations by visualizing the models' internals, e.g., activation maps in Convolutional Neural Networks, which can quickly become too complex when dealing with many neurons \citep{carter2019activation}. 
\emph{Counterfactual explanations} are a category of explanations that aims to answer the ``why'' question with ``because if it was \textit{something different}, it would be \textit{this other thing} instead''~\citep{guidotti2022counterfactual}. 
Counterfactuals are typically computationally intensive, and generating meaningful counterfactual examples depends on the task context \citep{artelt2019on}. 

Undoubtedly, explaining AI requires numerous considerations. 
Despite inherent limitations, each explanation technique enhances the explainability and transparency of AI systems, thereby advancing the overarching objective of fostering trustworthy and accountable AI applications. 
Many software libraries exist that make employing XAI methods straightforward. 
For Python, the libraries SHAP \citep{lundberg2017unified}, LIME \citep{ribeiro2016should}, Captum \citep{kokhlikyan2020captum}, and scikit-learn \citep{scikit-learn} are widespread and cover most XAI method categories. 
The DALEX \citep{biecek2018dalex} library offers model-agnostic explanations for the programming language R. 

\subsubsection{Methods to Evaluate Transparency and Explainability}
Evaluating explanation methods is vital for assessing their correctness, efficacy, and practical utility. Various approaches for estimating the effectiveness and quality of explanations have been introduced and can be divided into the following categories \citep{doshi2017towards}: 
\begin{itemize}
    \item \emph{Application-grounded evaluations} that involve human participants performing realistic tasks and offer insights into how XAI methods work in real-world scenarios.
    \item \emph{Human-grounded evaluations} that use simplified tasks for human participants to assess the comprehensibility and usefulness of explanations provided by the AI systems.
    \item \emph{Functionally-grounded evaluations} that rely on proxy tasks without human involvement, focusing on XAI algorithms' functionality and their scores against a pre-defined metric of interpretable quality. 
\end{itemize}

While application and human-grounded approaches focus on the plausibility and usefulness of explanations to users, functionally-grounded evaluations estimate the \emph{correctness} of XAI algorithms. 
Various properties of explanation methods can be examined to determine if they function correctly \citep{hedstrom2023quantus}. Some of these properties are: 
\begin{itemize}
    \item \emph{Faithfulness} \citep{alvarez2018towards,samek2016evaluating,vsimic2022perturbation} estimates how accurately explanation methods identify features in the input driving the model prediction.
    \item \emph{Robustness} \citep{montavon2018methods,alvarez2018robustness} measures an explanation method's sensitivity to input perturbations. 
\item \emph{Localization} \citep{Selvaraju_2017_ICCV,fong2017interpretable} identifies if the explanation method focuses correctly on the desired regions of interest. 
\item \emph{Complexity} \citep{bhatt2020,nguyen2020quantitative} measures the conciseness of explanations, where less complex explanations are deemed more interpretable than more complex ones.
\item \emph{Randomization} quantifies an explanation method's sensitivity to modifications of model parameters. 
\item \emph{Axioms} \citep{NEURIPS2018_294a8ed2,Kindermans2019} define criteria that an explanation method has to fulfil.
\end{itemize}

Hence, careful identification of evaluation aspects is necessary to address context-specific concerns, such as faithfulness, robustness, or comprehensibility. 
\rev{For a detailed overview of evaluation metrics for transparency and explainability, please also see~\citep{hulsen2023explainable}, in which metrics such as simulatability, decomposability, coherence, or comprehensiveness are mentioned.} 
Unfortunately, software libraries that offer metrics for validating explanation methods are scarce; among the few existing ones are Quantus \citep{hedstrom2023quantus} and AI Explainability 360 \citep{aix360-sept-2019}.

\subsubsection{Open Issues and Research Challenges}
The requirement for transparency and explainability of AI faces several open challenges. First and foremost, the research community needs to fully agree on a common, clear, and precise definition for transparency in AI systems, which currently leads to ambiguity regarding what explanations should entail. 
For instance, properly calibrating AI explanations to instill the correct amount of trust in AI models is crucial but complex \citep{wang2019designing}, as it requires a balance between providing understandable insights without oversimplifying or overwhelming users and, at the same time, without over or underselling the explained AI model's capabilities. 
Additionally, tailoring explanations for diverse user groups and individuals remains challenging, as different stakeholders require different explanations at varying levels of granularity and detail \citep{miller2017explainable,mendoza2023importance}. 
Furthermore, the evaluation of transparency and explainability of AI models is challenging, and developing intuitive user interfaces for explanations poses a design challenge, requiring informative yet user-friendly interfaces that follow ``XAI UI guidelines''~\citep{liao2020questioning,wolf2019explainability}. 

\rev{Finally, ensuring transparency and explainability in large language models and generative AI systems presents unique difficulties due to their complexity, and it is also unclear how their explanations should look like~\citep{schneider2024explainable}.}  

\subsection{Requirement 4: Robustness and Accuracy} \label{s:robustness}

\subsubsection{Definition of Robustness and Accuracy}
Robustness and accuracy are key properties of any AI system, and ensuring them is an essential part of the AI model development. Robustness and accuracy - in loose terms - refer to how ``adequate'' or ``correct'' the outputs of an AI model are. In contrast to other requirements - such as fairness or transparency - sufficient robustness and accuracy are required for \textit{any} AI model, independent of its specific purpose~\citep{huber2004robust}. 

AI training algorithms are typically designed for general problem settings (e.g., image-classification tasks). A specific AI model is then developed for a particular problem.
In many problem settings, various models can be employed. However, in complex settings, identifying a suitable model becomes challenging, often requiring the application of a model despite uncertainties regarding its suitability. 
In such a case, it is important to use models that produce reasonable results even if they are used in settings they were originally not designed for.
Under challenging conditions, some models may behave unpredictably and produce unstable outputs, whereas other models exhibit more constant behavior, and the quality of their outputs differs only slightly from the optimal setting. Clearly, it is preferable to use the latter class of models.
More concretely,~\citet{huber2004robust} outline three key properties an AI model should ideally possess. The model should: 1) achieve optimal or near-optimal results if it is applied in exactly the setting it was designed for; 2) degrade the quality of the results only slightly if it is subject to small deviations from the assumed setting; 3) not trigger nonsensical or dangerous outputs if it is applied in settings with large deviations from the assumed setting. 
An AI model is considered \textit{robust} if it meets properties 2) and 3), while property 1) is essential to ensure a sufficiently  \textit{accurate} model.

It is optimal to achieve high values in both, accuracy and robustness, but this is rarely possible. Instead, there is typically a trade-off between robustness and accuracy. While some models are more robust and can be applied across different settings, they come at the cost of lower accuracy. For example, a face detection algorithm could have very high accuracy in a highly specific setting, e.g., fixed camera type, fixed angle, and fixed lightning conditions, but might fail as soon as one of those parameters changes. A different model, on the other hand, might show slightly lower accuracy but perform similarly in various settings, e.g., different camera types and lighting conditions. 

Considering robustness and accuracy are crucial in all phases of the AI-lifecycle (see Section~\ref{sec:background}). In the design phase, it is important to make choices that do not compromise the accuracy of the model, for example, in the selection of appropriate training and testing data. The development phase is particularly essential, where one of the core tasks of every AI system development is to ensure these qualities. Finally, in the deployment phase, robustness and accuracy metrics need constant monitoring - especially in the case of continuously changing systems - to ensure a well-working AI model~\citep{hamon2020robustness}. 

\subsubsection{Methods to Establish Robustness and Accuracy}
Ensuring accuracy is the core of AI model development and part of all best practices. Additional core considerations are the appropriate choice of 1) target metrics the model is trained on, 2) data splitting techniques (e.g., train-test-splits), and 3) model selection methods. These three points are part of AI model developments that, in practice, are often done properly but not documented in a sufficient manner. To generate trust in AI models, it is essential to both deliver quality and document all relevant choices made in the process. These include, for example, the choice of suitable evaluation metrics, i.e., not only is it necessary to document the choice (e.g., ``F1-score''), but also the reasoning for that choice (e.g., ``classification problem with unbalanced data'')~\citep{huber2004robust,hamon2020robustness}.

Ensuring robustness can be done in two principal ways: 1) by restricting potential models to model types shown to be more robust (e.g., multilinear regression is generally more robust than deep learning), or 2) by explicitly evaluating model robustness and incorporating it in the model selection process. In the process of model selection, certain model types can be adapted to increase robustness. For example, fragile models can improve robustness by introducing mechanics that ignore certain data points or limit their effect, e.g., a drop-out layer in neural networks~\citep{krizhevsky2012imagenet}, or thresholds~\citep{kim2012robust}.
This has the advantage that the model learns to rely less on specific data points and focus more on the general information depicted in the majority of the data.
However, ``reserved'' data usage has a cost: The model has less data to work with, which puts it at a statistical disadvantage compared to fragile models that use all data. This cost is particularly high when the model is used in a setting where accuracy is more important than robustness~\citep{fisher1922mathematical}. 

\subsubsection{Methods to Evaluate Robustness and Accuracy}
In the AI literature, there are many different forms of robustness, e.g., robustness to domain shift \citep{blanchard2011generalizing,daomaingen2013,DBLP:journals/corr/abs-2007-01434}, adversarial robustness~\citep{nicolaeAdversarialRobustnessToolbox2019,xu2020adversarial}, robustness to noise \citep{zhu2004class,garciaEffectLabelNoise2015}, robustness to non-adversarial perturbations \citep{hendrycks2019benchmarking, rusak2020simple,scher2023testing} and others. While some generic robustness scores have been proposed \citep{wengEvaluatingRobustnessNeural2018a, sharmaCERTIFAICommonFramework2020a}, they do, in fact, only measure specific types of adversarial robustness.  Therefore, there is no single unified notion of robustness. Accordingly, as with most other aspects of trustworthy AI, the type of robustness to be considered depends on the context.  

Accuracy can be measured with a wide variety of metrics. In technical terms, accuracy is simply the fraction of correct outputs of an AI model~\citep{naidu2023review}. 
The goal of measuring accuracy is to measure how ``good'' or ``correct'' the outputs of the AI model are. How ``good'' or ``correct'' is defined in practice and highly depends on the application at hand. Therefore, there is no single generally applicable accuracy metric. The choice of which metric or metrics are applicable depends on the type of problem the AI model attempts to solve (e.g., regression, classification, ranking, translation), and on the particular properties of the application. For instance, for classification tasks with balanced classes, accuracy is a useful metric. However, classification tasks with highly imbalanced data, this can be misleading, and metrics such as precision and recall are more appropriate. Regression tasks require very different evaluation metrics than classification tasks. Examples are root-mean-square error or mean absolute error. Which one is more appropriate again depends on the application at hand. The same holds for ranking tasks, which require yet other types of metrics (such as mean reciprocal rank or mean average precision). An overview of these commonly used metrics can be found in \citet{poretschkin2023guideline}.

\subsubsection{Open Issues and Research Challenges}
\rev{The ongoing surge in generative AI models has opened a new challenge for existing models with respect to accuracy and robustness. It has been shown that generative models are, to a certain extent, capable of producing adversarial examples that cause catastrophic outputs in existing fragile AI models~\citep{han2023interpreting}. Moreover, some of these examples are transferable from one system to the next and can be re-used to cause failure in a number of different AI systems~\citep{wang2023towards}. Currently, there is no widely applicable easy strategy to address these issues. Existing fragile models need to be replaced with more robust models, and higher accuracy needs to be established for robust models, which are currently not able to compete with their fragile counterparts}.

Providing accuracy is the core of machine learning and AI, and thus, methods ensuring AI applications are accurate \rev{need to be} integrated with the development and improvement of the models. Accuracy evaluation metrics are well-established in statistics and machine learning, and their computation is generally straightforward. However, the choice of a proper metric and the definition of its thresholds are much more complex. While best practices exist, no formal guidelines are available. Also, despite a wide range of established accuracy metrics, there is a need for additional, new accuracy metrics that are specifically developed and tailored to the particularities of distinct applications~\citep{naidu2023review} \rev{-- and perhaps also tailored to permit robust solutions}. 

The field of robustness faces a variety of open issues and limitations. Robustness is a broad concept that current research strands do not necessarily cover in all aspects. In part, no specific methods - besides best-practice examples - are available for increasing robustness (e.g., robustness against noise). Especially the problem of adversarial attacks is constituted in a ``cat and mouse game'': if a specific attack strategy is known, AI systems can be made robust against it by incorporating the attack in the training procedure – known as adversarial training. However, this does not guarantee robustness towards a new attack that has not yet been part of the training.
Another open issue is caused by composite systems, where multiple AI components are combined or in situations where AI evaluation is part of a larger product/solution. Moreover, yet unanswered, is how to assess accuracy and robustness in evolving (learning) AI systems that are constantly updated (in some cases with every single user interaction)~\citep{hamon2020robustness}. 
In general, data quality, as well as model training and selection, are very important for AI systems, as these aspects influence accuracy and robustness, among other qualities such as fairness. Nonetheless, currently, no unified quality concept is available, even though basic automated tests are feasible.

\subsection{Requirement 5: Privacy and Security} \label{s:privacy}

\subsubsection{Definition of Privacy and Security}
Privacy and security are indispensable pillars supporting the trustworthiness and ethical use of AI systems. 
Privacy refers to the data used as input for the AI model and to protecting information that belongs to the data owner.
This information must not be disclosed to any third parties and may only be disclosed to parties that the data owner defines. 
Security, on the other hand, pertains to the AI model itself and is linked to defending it against any malicious attacks that aim to impact or manipulate it in an undesired or harmful way. Both privacy and security risks can arise along the whole life cycle of AI systems (see Section~\ref{sec:background}). 
Existing countermeasures span a broad spectrum, encompassing methods from manipulating the input data of AI models for ensuring privacy and security and designing AI models that are by themselves private and secure to recent advances that allow the protection of AI models during the inference process, i.e., during deployment~\citep{elliott2022ai}. 

If AI is utilized in critical areas such as healthcare, autonomous vehicles, or national security, it may even endanger human safety. 
Incidents like the unintended memorization of sensitive information by large language models\footnote{\url{https://incidentdatabase.ai/cite/357}} highlight the tangible privacy and security risks associated with AI. 
These examples serve as a reminder of AI models' potential to compromise privacy and security inadvertently. 
In response, regulatory bodies, particularly in the European Union, have been proactive in updating legal frameworks to address these challenges: The General Data Protection Regulation (GDPR) and the AI Act are prime examples of such regulatory efforts, aiming to establish clear guidelines for AI design, development, and deployment~\citep{zaeem2020effect}. 

\subsubsection{Methods to Establish Privacy and Security}
Understanding the weaknesses of AI systems and identifying the diverse kinds of attacks on privacy and security is critical for developing defense strategies and, consequently, evaluating their effectiveness. 
Attacks can be classified based on several aspects, including the attacker's capabilities and the attack goal.
For example, attackers can deviate from the agreed protocol (active/malicious) or try to learn as much as possible without violating the protocol (passive/semi-honest/honest-but-curious). 
Moreover, an attacker may be assumed to have finite or infinite computational power. 
Based on the attacker’s knowledge, one can differentiate between black-box attacks (which only access the model’s output), white-box attacks (which access the full model), and gray-box attacks (which gain partial access). 
In the following, we classify attacks based on the attack goal, i.e., evasion attacks, poisoning and backdoor attacks, and privacy attacks~\citep{bsi2023security,bae2018security}:

\begin{enumerate}
    \item \textit{Evasion attacks} (including adversarial attacks) aim to mislead AI models through carefully crafted inputs, forcing incorrect predictions. 
    \item \textit{Poisoning attacks} corrupt the training process, while \textit{backdoor attacks} insert hidden triggers into models. 
    \item \textit{Privacy attacks} seek to extract sensitive information from AI models. The most common privacy attacks include:
\begin{itemize}
    \item Membership inference attacks aim to determine whether a specific data sample was used in the training phase of the AI model.
    \item Attribute inference attacks aim to infer sensitive attributes, e.g., the gender of individual records.
    \item Model inversion attacks aim to infer features that characterize classes from the training data.
    \item Model extraction and stealing attacks aim to reconstruct the model’s behavior, architecture, and/or parameters. 
\end{itemize}
\end{enumerate}

~\citet{bsi2022security} list best practices to defend against the aforementioned attacks. In the following, we briefly provide examples for each class of attacks. 
Countermeasures against \textit{evasion attacks} include: 1) certification or verification of output bounds, i.e., utilizing certification methods to calculate guarantees on the output distribution to certify the AI model's robustness, 2) adversarial retraining, i.e., incorporating perturbed samples into the training process, 3) injection of randomness into training, i.e., using random transformations to protect against attacks, 4) use of more training data, i.e., enhancing adversarial robustness with larger and more diverse training datasets, 5) multi-objective optimization, i.e., not only optimizing for accuracy but balancing between adversarial robustness and task-specific accuracy, and 6) attack detection, i.e., implementing detection methods for malicious inputs. 
The risk of \textit{backdoor and poisoning attacks} can effectively be mitigated by the following strategies: 1) use of trusted sources, i.e., ensuring reliability and trustworthiness of data models; 2) random data augmentation, i.e., employing data augmentation techniques to mitigate the effect of poisoned samples; 3) use of an auxiliary pristine dataset, i.e., supporting training with trusted data to dilute the impact of poisoned samples; 4) attack detection, i.e., applying techniques to identify poisoned samples or models, including analysis of data distributions and model inspections; 5) model cleaning, i.e., utilizing methods like pruning, retraining, or differential privacy to eliminate the influence of triggers or poisoned data; and 6) adversarial training, i.e., adapting adversarial training approaches to counter poisoning attacks, enhancing model resilience.

Overall, defending the security of AI models against a variety of attacks involves a multifaceted approach that combines diverse techniques and practices, highlighting the need for AI practitioners to continuously assess and update their defense strategies.
Similarly, the development of privacy-enhancing technologies (PETs) has been instrumental in protecting AI models from \textbf{privacy attacks}.
The following (incomprehensive) list of PETs details the most important technologies, which currently form the forefront of research in private and secure AI computations.
\begin{itemize}
    \item \textit{Homomorphic encryption (HE)}~\citep{gentry2009fully,lncsRechbergerW22,iscSmart16,tifsPhongAHWM18} supports the performing of certain operations on encrypted data (i.e., without decrypting it). This allows privacy in cloud-based AI services to be maintained without exposing private data or model details. However, HE requires substantial computational resources and entails high computational costs. 
    \item \textit{Secure multi-party computation (MPC)}~\citep{lncsRechbergerW22,ftsecEvansKR18} allows collaborative computation without revealing individual inputs. In the context of AI, MPC is especially useful for collaborative learning and private classification, though it requires significant communication overhead for many participants. 
    \item \textit{Differential privacy (DP)}~\citep{tamcDwork08,fttcsDworkR14} 
     bounds the maximum amount of information that an AI model's output discloses about an individual data point by incorporating curated noise into the computation. 
    Specifically, noise can be added either to the input data, during the training process, or to the output~\citep{friedman2016differential}. 
    While effective in various AI applications, including deep learning~\citep{ccsAbadiCGMMT016} and recommender systems~\citep{mullner2023differential}, DP's main challenge is the trade-off between privacy protection and accuracy.
    \item \textit{Federated learning (FL)}~\citep{kbsZhangXBYLG21,spmLiSTS20} is a machine learning approach that allows multiple clients, like mobile phones, to collaboratively learn a model by training locally and sharing updates with a central server. This method enhances privacy by keeping data local, although there is a risk of data reconstruction from model updates~\citep{cvprYinMVAKM21,nasr2019comprehensive,ren2022grnn}. 
    \item \textit{Synthetic data}~\citep{slokom2018comparing,liu2022privacy} mimics real data's statistical features to enable the AI model to still learn the real data's features, but without using the real data.
    This offers a way to preserve privacy in data sharing, yet it is not immune to reconstruction risks~\citep{ussStadlerOT22}. 
    \item \textit{Transfer learning}~\citep{zhuang2020comprehensive}, while not a PET per se, contributes to privacy by fine-tuning pre-trained models on new tasks with minimal data, reducing the need for large private datasets~\citep{gao2019privacy}. Similar ideas are also employed by PETs based on meta-learning~\citep{muellner2021robustness}. 
\end{itemize}

The described defense methods can also be combined to increase privacy and security. 
For example, DP can mitigate the risk of reconstruction in FL~\citep{tifsWeiLDMYFJQP20} and synthetic data~\citep{prdcTaiLHW22,ussStadlerOT22}.

\subsubsection{Methods to Evaluate Privacy and Security}
The vulnerability of AI models to privacy and security attacks can be assessed using two complementary approaches: mathematical analysis and attack-based evaluation. 
\textit{Mathematical analysis} offers formal proofs of privacy and security features within a system, much like cryptography, guaranteeing system security under certain assumptions (e.g., DP). 
This method is crucial, especially when introducing new privacy or security techniques, as it requires thorough checks for implementation errors and the appropriate selection of parameters.
On the other hand, \textit{attack-based evaluation} gives us practical insight into how an AI model reacts to various attack strategies.
This method tests the model's vulnerability to different attacks and determines its resilience by using various metrics~\citep{wagner2018technical,pendleton2016survey}. 
These metrics might include the attacks' success rate, the effort required to breach the model (measured in iterations), the precision of the attack, and the smallest necessary data alterations to compromise the model successfully~\citep{bsi2023security}. 
The choice of metrics depends on the nature of the attack and on assumptions about the attacker's skills and knowledge. It is tailored to each specific scenario and model based on potential threats and existing literature.
However, it is important to acknowledge the limitations of attack-based evaluations. 
While they can pinpoint specific weaknesses and vulnerabilities, they do not offer a comprehensive guarantee of privacy or security. 
Additionally, these evaluations only cover known attack scenarios, leaving the potential for undetected vulnerabilities against new or complex attack techniques.

\subsubsection{Open Issues and Research Challenges}
Despite the existing countermeasures, the AI privacy and security field still faces numerous unresolved challenges. 
Many defense strategies cannot fully mitigate the models' vulnerability to attacks, especially not to adversarial and poisoning attacks. 
\rev{Additional challenges emerge with the increasing advancement of generative AI, particularly in models that rely heavily on unstructured data such as text. 
For example, establishing clear boundaries on what constitutes private information becomes increasingly difficult due to the inherent complexities of unstructured data \citep{brown2022does}.}
PETs often introduce trade-offs, such as increased computational demands (HE and MPC)~\citep{moore2014practical}, reduced prediction accuracy and increased unfairness (DP)~\citep{ccsAbadiCGMMT016,bagdasaryan2019differential,mullner2024impact}, or a surge in communication overhead while having no privacy guarantees (FL)~\citep{ALMANIFI2023100742,bagdasaryan2020backdoor}. 
Therefore, integrating PETs smoothly into AI systems without compromising performance remains complex and requires further research.
\rev{Just} as for fairness and robustness, evaluating privacy and security when combining multiple AI components is challenging. Adding components that protect against one identified risk can even introduce new vulnerabilities~\citep{debenedetti2023privacy}.

In general, fostering secure model sharing and privacy-preserving collaboration, developing standardized evaluation metrics, and preparing for advanced AI threats necessitate a collaborative approach among researchers, developers, and policymakers. 
Ongoing research and shared best practices will be crucial for building a secure, privacy-conscious AI ecosystem.

\subsection{Requirement 6: Accountability} \label{s:accountability}

\subsubsection{Definition of Accountability}
Another key requirement for trustworthy AI is accountability. At its heart, accountability is the obligation to notify an authority of one's conduct and to justify it~\citep{Bovens2007account,Brandsma2012Cube,Novelli2023Account, Hauer2023inspec,wieringa2020literatureAI}, whereas responsibility includes explicit obligations defined in advance~\citep{Bivins2006ResponsibilityAA} and can be seen as a subcategory of accountability~\citep{leon2021account}. Liability is closely related to accountability and means legal responsibility, including sanctions for misbehavior. In this article, we, therefore, see liability as a sub-concept of accountability and solely use the term accountability.  

From a conceptual perspective, accountability can also be defined as a virtue or as a mechanism~\citep{Bovens2010Concepts}. 
Accountability ``refers to the idea that one is responsible for their action – and as a corollary their consequences – and must be able to explain their aims, motivations, and reasons''\footnote{\url{https://digital-strategy.ec.europa.eu/en/library/ethics-guidelines-trustworthy-ai}}.  

The definition of \citet{Bovens2007account} is widely used as the basis for addressing accountability and identifies the following key elements of accountability: actor, forum, relationship between these two, account, and consequences. The actors, as natural persons, groups or organizations (e.g., developers, deployers, manufacturers, or users of AI systems), shall be able to explain their actions (e.g., used models and data, intended use, planned outcomes, and potential malfunctions of AI systems) by certain criteria to the forum (e.g., a court, a supervisor, an auditor), that can ``pose questions and pass judgments''. The relationship between actor and forum can vary and involves individual, hierarchical, collective and corporate accountability. Finally, there will be consequences (e.g., fines for non-compliance with rules). 

\subsubsection{Methods to Establish Accountability} 
When addressing accountability features such as context, the range of actions taken, the acting entity, the forum as the bearer of interests and imposed standards, processes, and implications must be considered \citep{Bovens2007account} to be able to achieve compliance, report, oversight, and enforcement~\citep{Novelli2023Account}.
Thus, accountability is always relational~\citep{Bovens2007account}, contextual \citep{Lewis2020StandardizationAI} and involves single persons, other entities, as well as groups and societies. Depending on impacts, different levels of accountability are required~\citep{Chech2021algorithmicaccount}.

Accountability systems range from hard law regulations over functional roles within organizations \citep{Novelli2023Account} to social norms that, in turn, form the basis for decision-making and behavior~\citep{leon2021account}. 
In the field of software development, responsibility involves maintaining quality in the design process~\citep{eriksen2002designaccount}, implementing tools for characterizing system failure \citep{Nushi2018AI}, as well as using transparency and inspection mechanisms \citep{Hauer2023inspec}. So called ``algorithmic accountability'' is also described as the expectation that people along the AI-lifecycle (see Section~\ref{sec:background}) will comply with legislation and standards to ensure the proper and safe use of AI and involves not only the use, design, implementation, and consequences but the whole ``socio-technical process'' \citep{Hauer2023inspec,Novelli2023Account,wieringa2020literatureAI}. 
Thus, accountability shall ensure compliance with requirements such as fairness, transparency, and robustness~\citep{FloridiMeta2022,Novelli2023Account}. Therefore, it also requires that mechanisms for auditability, minimization, and reporting of negative impacts, trade-offs, and redress are in place. Therefore, accountability must be ensured along the whole AI-lifecycle - in the design phase, the development phase, and the deployment phase.

\subsubsection{Methods to Evaluate Accountability} 
Due to the versatility of accountability, its evaluation is challenging. Numerous approaches for evaluating accountability regarding AI systems are put forth\footnote{In this article, we only present individual approaches as examples without any claim to completeness.}. For example, \citet{Tagiou2019toolframework} suggest a ``a tool-supported framework for the assessment of algorithmic accountability'' that focuses on both algorithmic and organizational aspects and \citet{Chech2021algorithmicaccount} proposes the ``\textit{Accountability Agency Framework (A3)}'' as an analytic lens as a qualitative, explorative, and complementary tool to assess algorithmic accountability, which is based on \citet{Bovens2007account}'s definition of accountability. Their framework encompasses four steps: requesting information, providing account, imposing consequences, and effective change. Additionally, it provides a series of guiding questions for assessing algorithmic accountability~\citep{Chech2021algorithmicaccount}. \citet{Xiametrics} proposed a granular AI Metrics catalog that includes process, resource, and product metrics and is specially designed for generative AI.  
Besides, numerous other, mainly contextualized frameworks, which range from accountability in organizations \citep{Buhmann2019manage}, public reason \citep{Binns2018PublicReason} and public service \citep{Brown2019Publicservice}  to “AI robots accountability” \citep{Toth2022Robotics} frameworks have been proposed. From a qualitative perspective, approaches that, for instance, take human rights into account are discussed \citep{McGregor2019HumanRightsacccount}.  

In general, \citet{Brandsma2012Cube} suggest a so-called ``\textit{accountability cube}'' as a quantitative assessment tool for assessing accountability, considering three dimensions of accountability processes: information, discussion, and consequences/sanctions. Accordingly, accountability is “high” if there is much information, intensive discussions, and several opportunities to impose consequences. This approach can be applied in various contexts along the AI-lifecycle. Without any closer examination of the approach itself, we use the accountability cube to exemplify possible evaluation criteria of algorithmic accountability. 

To start with, much information is given if people are aware of basic technical outlines, chances, and risks of AI in the respective context and know their obligations along the lifecycle of AI, including, for instance, information, documentation, and risk-assessment obligations. From our point of view, AI literacy is essential to this. Discussion is intensive if an informed exchange of views on AI systems and regulation – whether formal or informal – takes place between multiple stakeholders (e.g., policymakers, NGOs, technical experts, civil society, as well as companies and individuals). Besides, there shall be meaningful opportunities to explain actions (e.g., using certain design concepts/training data or using AI systems in certain situations). Finally, effective and proportionate consequences (e.g., penalties for non-compliance with rules and effective redress) shall be in scope. This, in turn, creates a need for clear and feasible rules. 

Notably, the weight of these principles can vary. To exemplify, the “accountability rate” might still be high if there are clear non-binding standards with no legal consequences that are widely adhered whereas it is lower if there are binding rules that are not being followed due to societal rejection or inefficient enforcement. The weight of the principles might also vary in different contexts. For instance, in policymaking, intensive discussion might have a higher priority than in company internal processes. 

\subsubsection{Open Issues and Research Challenges}
In practice, evaluating algorithmic accountability poses severe problems. One of the biggest challenges is that algorithmic accountability is a ``multifaceted and context-sensitive challenge'' \citep{Chech2021algorithmicaccount}. At present, standards, standardized methods, and metrics covering different aspects of the AI-lifecycle, from design to deployment, are still incomplete and, therefore, do not provide sufficient legal security. Vague terms confront norm addressees with legal uncertainty when interpreting these norms. In turn, organizations are unable to implement sufficient accountability mechanisms within their organization.  

On the one hand, accountability gaps arise if rules are inconsistent, unclear, or not feasible, and therefore, they lead to ineffective redress of victims. On the other hand, rules which are too strict generate accountability surpluses, which in turn decrease technological and economic growth~\citep{Bovens2007account, Novelli2023Account}. AI policymakers aim to close these accountability gaps that might arise due to the unpredictable, opaque nature of AI systems \citep{Novelli2023Account,Busuioc2020Account}. Several measures, like model certification, algorithmic impact assessments, real-world testing, and third-party audits, could foster accountability \citep{Busuioc2020Account}. Such measures are also included in the AI Act. For example, there are documentation, information, and transparency obligations for providers of high-risk AI systems, there are third-party checks, and testing in real-world laboratories is enabled. Rules are also amendable according to technical changes, demonstrating effective change if needed.  

Notably, developing sufficient rules, including ethical and technical standards, that cover the whole AI-lifecycle is challenging, as AI systems are complex and based on various programming methods, developing rapidly and can have wide-ranging effects on people \citep{Chech2021algorithmicaccount}. \rev{Generative AI systems seem to exacerbate this problem due to their large scale, complexity, and adaptability~\citep{Xiametrics}. Consequently, it is particularly difficult to find suitable metrics for evaluating the trustworthiness of generative AI.} To ensure ``actionable'' accountability, both technical and non-technical aspects, among them legal and ethical aspects, must be considered \citep{Stix2021AIpolicy}.  When creating rules and standards on AI, it is crucial to weigh up technical and economic aspects. An informed dialogue between policymakers, (technical) experts, and civil society is essential to reaching sufficient rules and avoiding unnecessary bureaucracy.  

\section{Conclusion and Future Research}
\label{sec:conclusion}

In this paper, we investigated the following six requirements of trustworthy AI: 1) human agency and oversight, 2) fairness and non-discrimination, 3) transparency and explainability, 4) robustness and accuracy, 5) privacy and security, and 6) accountability. \rev{With respect to our guiding research question introduced in Section~\ref{s:intro} (i.e., \textit{What is the current state of research regarding the establishment and evaluation of comprehensive - technical, human-centered, and legal - requirements of trustworthy AI?}), our findings confirm that ensuring AI systems meet these criteria is a complex endeavor requiring technical solutions, policy frameworks, and interdisciplinary collaboration. Additionally, our article demonstrates that while evaluation and validation methodologies for technical requirements, such as robustness, can often rely on well-established metrics and testing procedures (e.g., model accuracy), assessing human-centric considerations demands more nuanced approaches that take into account ethical, legal, and cultural factors. Therefore, we believe that our article complements existing surveys and assessment lists (e.g., ALTAI~\citep{ALTAI,radclyffe2023assessment}) of trustworthy AI.} 

\begin{figure}[t!]
\begin{center}
\includegraphics[width=0.70\linewidth]{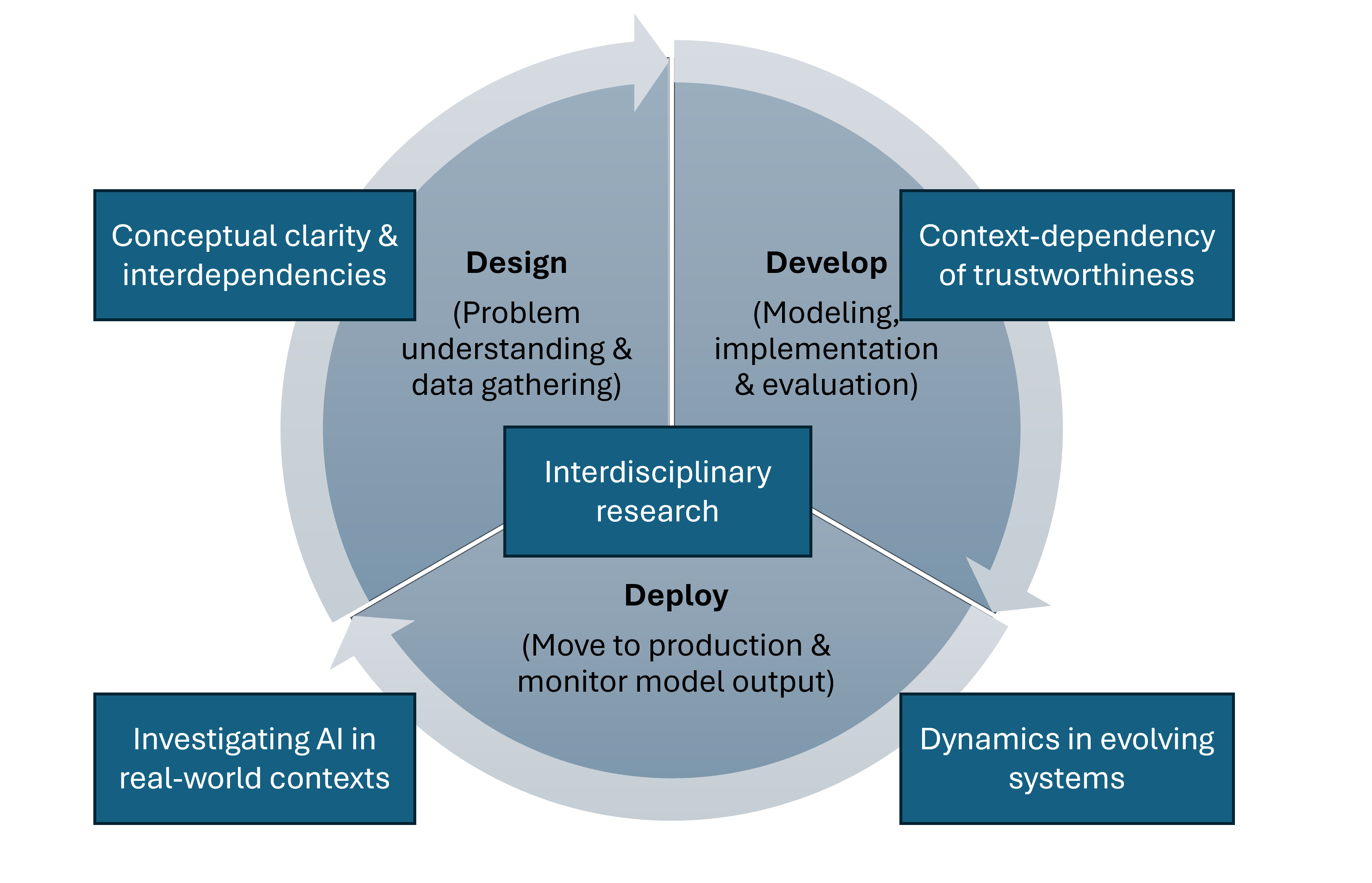}
    \caption{Overarching research challenges identified in this paper in relation to the AI-lifecycle phases.} 
    \label{fig:issues}
\end{center}
\end{figure}

This section further synthesizes our key observations across these very different aspects of AI systems in relation to their trustworthiness and discusses the implications of this overarching analysis. 
Additionally, Figure~\ref{fig:issues} visualizes these overarching research challenges in relation to the phases of the AI-lifecycle mentioned in Section~\ref{sec:background}. 

\begin{itemize}
    \item \textbf{Interdisciplinary research.} 
    The interdisciplinary nature of trustworthy AI research becomes apparent when considering the different scientific foundations necessary to discuss the design, development, and deployment of trustworthy AI. This demand for interdisciplinarity is also recognized by initiatives like the human-centered AI (HCAI) workshops and sessions at AI conferences such as NeurIPS\footnote{\url{https://hcai-at-neurips.github.io/site/}}, as well as the FAccT community's work on fairness, accountability, and transparency in AI\footnote{\url{https://facctconference.org/}}. However, broader collaboration is needed. In particular, insights from social scientists, ethicists, and policymakers can complement technical research, for example, in fairness definitions, explainability, or human oversight. At the same time, interdisciplinary research comes with its own challenges, such as distinct disciplinary-specific jargon. Thus, agreeing on standardized, cross-disciplinary terminology remains an ongoing challenge in various subjects.
    \item \textbf{Conceptual clarity and interdependencies.} 
    Across all trustworthy AI requirements, we see the need to sharpen definitions and to consider interdependencies and relationships between concepts. This involves understanding the potential trade-offs between requirements, such as fairness and accuracy or explainability and privacy. Such conceptual clarity and knowledge of interdependencies will help in designing trustworthy AI with regard to specific requirements while allowing for informed discussions of trade-offs. Therefore, it is essential to consider potential trade-offs and interdependencies already when designing trustworthy AI systems. 
    \item \textbf{Context-dependency of trustworthiness.} 
    Our research indicates that AI requirements are very context-dependent. This means that any insights for developing trustworthy AI are challenging to transfer across different contexts due to cultural and application-specific aspects. Different interaction patterns between humans and AI will be appropriate in different contexts, and definitions of trustworthiness vary between societies and applications. This raises questions about the sufficiency of existing evaluation frameworks and suggests the need for new approaches that can better adapt to contextual differences. 
    Additionally, if an audited algorithm is reused and fails to meet requirements in a different context, assigning responsibility becomes complex. Even more difficult is the handling of AI solutions that consist of multiple interacting components, i.e., composite systems. Understanding how different components interact and affect each other is crucial when algorithms are reused in conjunction with other components. Even if single components are considered trustworthy, the results of their interplay potentially violate the requirements of trustworthy AI. Such uncertainties affect the licensing and use of software frameworks, which emphasizes the importance of developing licensing models that clearly outline accountability while promoting the responsible use of AI. 
    \item \textbf{Dynamics in evolving systems.} 
    One of the emerging issues in trustworthy AI is the potential of learning unintended facets during deployment. In evolving systems (i.e., systems that learn during deployment), in particular, the capturing of biases may lead to trustworthiness issues, especially with respect to fairness and non-discrimination. Such biases are often cognitive biases of users, which are acquired through ongoing learning cycles and require more sophisticated research to form a deeper understanding of related patterns and furthermore, develop approaches for detection and mitigation. This concern also highlights the necessity of dynamic and adaptive evaluation and simulation frameworks. Since the majority of trustworthy AI evaluation schemes operate in a static manner, additional research is needed to investigate, monitor, and capture long-term dynamics of trustworthiness.
    \item \textbf{Investigating trustworthy AI in real-world contexts.} 
    Due to the complexity of AI systems and their contextual dependencies, it is crucial to study their functionality in real-life contexts to gain a deeper understanding of their impact. The involvement of human factors, such as how a system is used by different people and how this fits into a complex socio-technical context, makes real-world investigations very challenging from a methodological standpoint. However, for some requirements on AI systems, such as fairness or human agency, this may be particularly important to the extent that fully valid statements about these concepts may only be made after investigation in real-world contexts. Thus, monitoring the trustworthiness of AI is an ongoing investigation, especially after the system has been deployed in a real-world context.   
\end{itemize}

\rev{Finally, and as outlined in Sections~\ref{s:human_agency}--\ref{s:accountability}, the current developments around generative AI and LLMs introduce new challenges for establishing and evaluating trustworthy AI. Therefore, future research also needs to investigate how existing trustworthy AI methods and definitions (e.g., fairness metrics for binary classification problems) can be transformed into more general settings provided by generative AI and LLMs.} 
We hope that our paper provides a reference point for both researchers and practitioners in the field of trustworthy AI and a starting point for future research directions addressing the open research challenges identified in this work and discussed in this section. 

\section*{Conflict of Interest Statement}
\rev{Author Tomislav Nad is employed by SGS Digital Services GmbH.} The remaining authors declare that the research was conducted in the absence of any commercial or financial relationships that could be construed as a potential conflict of interest 

\section*{Author Contributions}
DK and SK: literature analysis and writing of ``Fairness and Non-Discrimination'' section; creation of all figures; conceptualization, organization, writing, supervision, \rev{revision,} and submission of the paper. SS, VPS, PM, RK, and TN: composition and revision of the paper. KW: literature analysis and writing of ``Accountability'' section. LD, PM, AT, and TN: literature analysis and writing of ``Privacy and Security'' section. AF, EV, and VPS: literature analysis and writing of ``Human Agency and Oversight'' section. SS and MT: literature analysis and writing of ``Robustness and Accuracy'' section. IGME, IS, and VS: literature analysis and writing of ``Transparency and Explainability'' section.

\section*{Funding}
The work received funding from the TU Graz Open Access Publishing Fund.

\section*{Acknowledgments}
Parts of this manuscript are based on white papers created by SGS Digital Trust Services GmbH and Know-Center, which are available online via Zenodo: \url{https://zenodo.org/records/11207961}. 
Know-Center is a COMET Center within the COMET – Competence Centers for Excellent Technologies Program and funded by BMK, BMAW, as well as the co-financing provinces Styria, Vienna and Tyrol. COMET is managed by FFG.

\bibliographystyle{Frontiers-Harvard}
\bibliography{references.bib}

\end{document}